\documentclass[11pt,a4paper]{article}
\usepackage[hyperref]{acl2017}
\usepackage{times}
\usepackage{latexsym}
\usepackage{subfigure}
\usepackage{graphicx,epsfig}
\usepackage{multirow}
\usepackage{booktabs}
\usepackage{url}
\usepackage{amsmath}
\usepackage{amsfonts}
\usepackage{amssymb}

\usepackage{color}
\definecolor{redcolor}{rgb}{1.0,0.,0.}

\aclfinalcopy % Uncomment this line for the final submission

\title{On the ``Calligraphy'' of Books}

{\author{Vanessa Queiroz Marinho$^\ast$ \qquad 
Henrique Ferraz de Arruda$^\ast$ \qquad 
Thales Sinelli Lima$^\ast$ \\
{\bf Luciano da Fontoura Costa}$^\dagger$ \qquad 
{\bf Diego Raphael Amancio}$^\ast$ \\
$^\ast$Institute of Mathematics and Computer Science, University of S\~{a}o Paulo, S\~{a}o Carlos, S\~{a}o Paulo, Brazil \\
$^\dagger$S\~{a}o Carlos Institute of Physics, University of S\~{a}o Paulo, S\~{a}o Carlos, S\~{a}o Paulo, Brazil \\
{\tt \{vanessa.qm.1, h.f.arruda, thalessinelli, diegoraphael, ldfcosta\}@gmail.com} % \mla{complete}
}

\date{}

\begin{document}
\maketitle
%up to 200 words
\begin{abstract}
Authorship attribution is a natural language processing task that has been widely studied, often by considering small order statistics. In this paper, we explore a complex network approach to assign the authorship of texts based on their mesoscopic representation, in an attempt to capture the flow of the narrative.  Indeed, as reported in this work, such an approach allowed the identification of the dominant narrative structure of the studied authors.  
This has been achieved due to the ability of the mesoscopic approach to take into account relationships between different, not necessarily adjacent, parts of the text, which is able to capture the story flow. The potential of the proposed approach has been illustrated through principal component analysis, a comparison with the chance baseline method, and network visualization. Such visualizations reveal individual characteristics of the authors, which can be understood as a kind of calligraphy.
\end{abstract}

\section{Introduction}
The ever increasing availability of public content on the Internet -- including books, tweets, and blog posts -- has implied in many new  developments in several natural language processing (NLP) areas such as machine translation, sentiment analysis, and authorship attribution. Recently, advancements in the latter task have been achieved by using complex networks~\cite{Antiqueira2006a, Amancio2011a,Lahiri,Marinho2016BRACIS,AkimushkinAO16}.  

The network models used in many of these works are based on word co-occurrence. In this approach, each distinct word is represented by a node, and edges connect adjacent words. Although this networked representation has proven successful in many tasks, it is not without its share of problems. Co-occurrence networks do not portray the topical structure found in many texts and are usually devoid of community structure~\cite{topicseg}. In order to overcome this disadvantage, some techniques have been devoted to the \emph{mesoscopic} representation of texts~\cite{topicseg,de2017mesoscopic}. \citet{de2017mesoscopic} proposed a novel networked model, in which each node represents a respective set of consecutive paragraphs, while weighted edges express the similarity between nodes. Their proposed network is able to extract the organization and flow of text by effectively capturing the similarity between the blocks of text. {In addition, their method was employed to distinguish between real and shuffled texts. However, mesoscopic networks have not been applied to tackle other NLP tasks.}

Most researchers in the field of authorship attribution assume that each author has a signature (known as authorial fingerprint) that distinguishes his/her writing from the others~\cite{Juola:2006}. So inspired, we decided to test the hypothesis that these authorial fingerprints are also visible at a mesoscopic scale. { At this scale, distinctive graphical patterns of the course of the text emerge, akin to a ``discourse calligraphy'' of the author}. {Thus, in order to classify texts according to their authorship, we created mesoscopic networks from texts and employed a set of topological measurements.} In particular, the main goal of this paper is to probe whether the authors' writing styles correlate with the story flow of their books.

This paper is structured as follows: Section~\ref{sec:related_work} briefly describes the problem and some complex network approaches for authorship attribution. The process to create mesoscopic networks is explained in Section~\ref{sec:methods}. In addition, we also describe the dataset, the selected measurements and the machine learning algorithms in Section~\ref{sec:methods}. The obtained results are reported in Section~\ref{sec:results}. Finally, Section~\ref{sec:conclusion} outlines our conclusions and prospects for future work.

\section{Related Work}\label{sec:related_work}

Authorship attribution methods attempt to find the most likely author of a document~\cite{Stamatatos}. Since the seminal work conducted by~\citet{Mosteller}, authorship attribution has been a widely studied problem and several different approaches have been proposed. One of the first approaches consisted in analyzing the frequency of common words, such as \emph{to}  or \emph{the}, in order to classify political essays according to their authorship~\cite{Mosteller}.

Since then,~\citet{Mosteller}'s method has been enhanced to incorporate different attributes capable of qualifying writing styles. These include lexical, character, syntactic, and semantic features~\cite{Stamatatos}. Simple lexical and character features (e.g. frequency and burstiness of words and characters, average lengths of texts, and others) have been used in several works, as reported by~\citet{grieve2007},~\citet{Koppel:2009}, and~\citet{Stamatatos}. Most of these works have achieved good results by using, for example, the frequency of stopwords. Examples of syntactic information include the frequencies of POS tags and constituency-based parsing tree rules~\cite{Baayen,gamon:2004,Hirst2007}. Finally, semantic features can be extracted from semantic dependency graphs and from the semantic roles associated with some words~\cite{gamon:2004,Argamon2007}.

The usage of network analysis in authorship attribution has already been studied from different perspectives. \citet{Antiqueira2006a}, one of the first works in the area, extracted some measurements from co-occurrence networks and discovered that these could be used to characterize the writing style of authors. \citet{Amancio2011a} combined network measurements with the distribution of words to characterize the authorship of several books. \citet{Lahiri} carried out an in-depth authorship attribution study using more than 100 features extracted from co-occurrence networks. They found that local features (those extracted from individual nodes) outperform global features in the authorship attribution problem. 

Apart from using traditional network measurements, the frequency of network motifs involving three nodes~\cite{Milo} was found useful to characterize the writing style~\cite{Marinho2016BRACIS}. Instead of considering the text as a static structure, ~\citet{AkimushkinAO16} studied the topology evolution of co-occurrence networks extracted from different sections of the text. Unlike most of the previous mentioned works, in which stopwords are usually removed, ~\citet{segarra2013authorship} proposed an authorship attribution method based on networks formed only by stopwords.
 
\section{Methods}\label{sec:methods}
In this section, we describe the process to create mesoscopic networks from raw texts. We also detail the network measurements and machine learning methods.

\subsection{Mesoscopic Approach}
\begin{figure*}[!htpb]
 \centering
   \includegraphics[width=1.0 \linewidth]{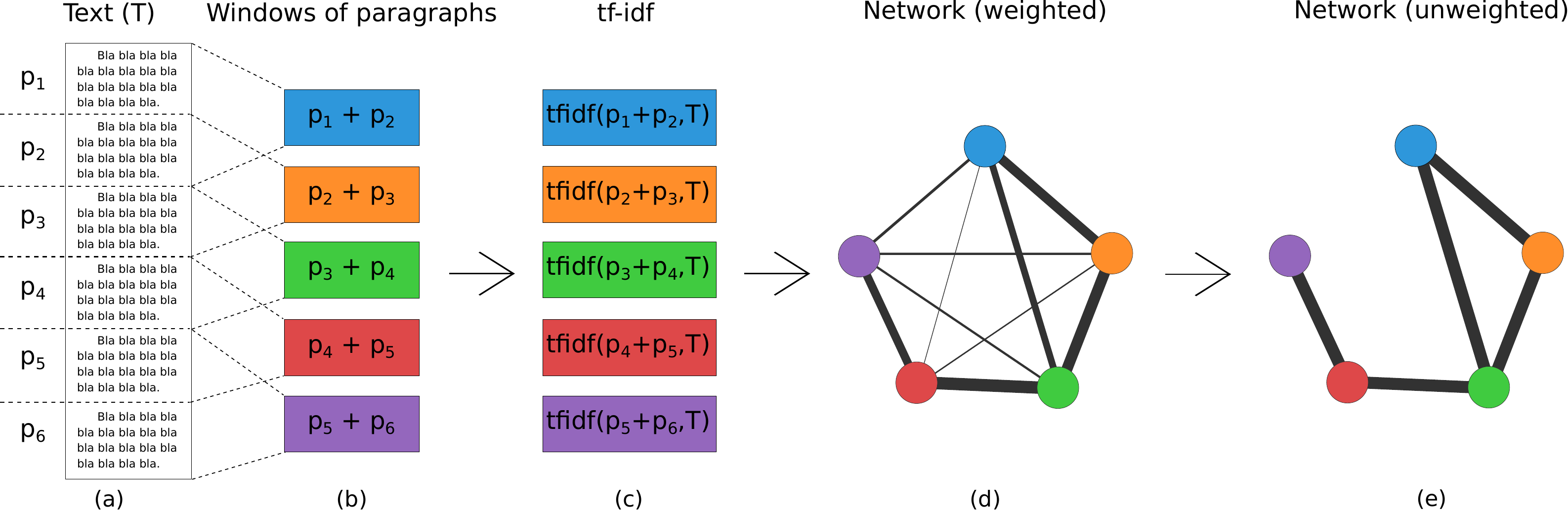}
    \caption{Illustration of the mesoscopic approach proposed by~\citet{de2017mesoscopic}. First, the text $T$ is divided into subsequent paragraphs (a). Overlapping windows with $\Delta = 2$ paragraphs are shown in (b). Then, the tf-idf map is computed for all windows (c). Each pair of nodes (windows) $i$ and $j$ is now connected by an edge, weighted by the cosine similarity between their respective tf-idf maps (d). Next, in the network pruning phase, the edges with the lowest weights are removed until the network reaches a given average degree $\left\langle k \right \rangle$. The network in (e) illustrates the obtained unweighted mesoscopic network with $\left\langle k \right \rangle = 2$.
}\label{fig:multiScale}
\end{figure*}
  
There are several ways to represent texts as complex networks, such as co-occurrence, syntactic, semantic or similarity networks~\cite{Mihalcea,cong2014approaching}. In this study, we adopt the mesoscopic network approach proposed by~\citet{de2017mesoscopic}. Such networks are able to represent the text unfolding along time, which is normally overlooked by traditional approaches. Moreover, these networks were used to classify documents between real and shuffled texts, using only simple statistics. The high accuracy rate obtained in that classification task led us to infer that mesoscopic networks are able to represent structural aspects of real texts, such as the organization and development of the author's idea.

In order to create the network from a given text ($T$), some preprocessing steps can be applied. In our study, we removed the stopwords, and the remaining words were lemmatized.  Figure~\ref{fig:multiScale} illustrates the methodology used to create mesoscopic networks. In the first step, shown in Figure~\ref{fig:multiScale}(a), the text is partitioned into a set of paragraphs, $T = (p_0,p_1,p_2,\cdots)$, where $p_i$ is a sequence of the preprocessed words belonging to the same paragraph $i$. Different from the co-occurrence networks, where nodes represent words,  in mesoscopic networks nodes encompass sequences of $\Delta$ consecutive paragraphs. More specifically, each possible subsequent set with $\Delta$ paragraphs, $W_i^\Delta = (p_i,p_{i+1},\cdots,p_{i+\Delta - 1})$,  represents a network node, as shown in Figure~\ref{fig:multiScale}(b). 

So as to account for the  importance of the words in a given paragraph, we applied the \emph{tf-idf}~\cite{Manning:1999} statistics, which was originally proposed to quantify the importance of a given word $w$ in a document $d$ given a corpus $D$. A $\text{tf-idf}(w,d,D)$ map is computed as
\begin{equation}
  \text{tf-idf}(w,d,D) = \frac{f_{w,d}}{n} \times \log\Bigg{(} \frac{|D|}{d_w}\Bigg{)},
\end{equation}
where $f_{w,d}$ is the frequency of word $w$ in the document $d$, $n$ is the total number of words in the document $d$, $|D|$ represents the total number of documents and $d_w$ is the number of documents in which $w$ occurs at least once. In order to apply the tf-idf measurement, we considered all the possible windows of subsequent paragraphs, $W_i^\Delta$, as the set of documents $D$ (see Figure~\ref{fig:multiScale}(c)). Finally, for each pair of nodes $i$ and $j$, a respective edge is created and its weight is calculated according to the cosine similarity between $\text{tf-idf}(W_i^\Delta,T)$ and $\text{tf-idf}(W_j^\Delta,T)$, {where $\text{tf-idf}(W_i^\Delta,T)$ is a tf-idf vector of all words, computed from a given set of paragraphs $W_i^\Delta$}. This step is illustrated in Figure~\ref{fig:multiScale}(d).

In order to convert the network from weighted to unweighted, the edges with the lowest weights can be removed, as described in Section~\ref{sec:pruning}. It should be noted that edges originating from adjacent paragraphs tend to have higher weights because of the implied overlap. Figure~\ref{fig:multiScale}(e) shows an example of unweighted network. In our experiment, we set $\Delta=20$, as empirically determined elsewhere~\cite{de2017mesoscopic}.
\subsection{Network Pruning}\label{sec:pruning}

Mesoscopic networks are complete weighted graphs, i.e. every node is connected to every other node~\cite{Newman2010}. In this paper, we repeatedly removed the edges with the lowest weights until each network reached a fixed network average degree $\left\langle k \right \rangle$. The average degree of a network $g$, with $E$ edges and $N$ nodes, is defined as
\begin{equation}
\left\langle k \right \rangle = \frac{2*E}{N}.
\end{equation}
 We used several values of $\left\langle k \right \rangle$, ranging from 5 to 50, by steps of 5.

\subsection{Network Measurements}
The following network measurements were extracted from the networks\footnote{For most of these measurements, we used the Igraph software package~\cite{igraph}}. Most of these measurements {(apart from assortativity)} apply to a single node.  So, in order to obtain more global characterization, we calculated the average, standard deviation and skewness (third moment) of each distribution. {The obtained statistics from these distributions were then used as features in the machine learning methods.}

\emph{\textbf{Degree}}: The degree quantifies the number of connections of a node~\cite{costa2007characterization}. Even though the average degree of all networks is the same as a consequence of network pruning, the degree of each node may still vary inside the network. Therefore, we used the standard deviation and skewness of this measurement, disregarding the average.

\emph{\textbf{Average Degree of Neighbors}}: The average degree of neighbors~\cite{pastor2001dynamical} quantifies how well connected are the neighbors of a node. 

\emph{\textbf{Assortativity}}: As described by~\citet{newman2003mixing}, the assortativity quantifies how likely it is for a given node to connect to other nodes with similar degree. Lower than zero values of assortativity are obtained when a node tends to connect to others with very different degrees.  When a node connects only to others with the same degree, the assortativity becomes one. Null assortativity indicates that there is no correlation. 

\emph{\textbf{Clustering Coefficient}}: This measurement reflects how well interconnected are the neighbors of a given node~\cite{watts1998collective}. 

\emph{\textbf{Accessibility} ($h = \{2,3\} $)}: The accessibility of a node $i$ is based on Shannon's entropy~\cite{Shannon:1963} of the probability of accessing nodes at the \textit{$h^{th}$} concentric level, centered at $i$, by a given dynamics starting at that node~\cite{travenccolo2008accessibility}.  Here, we adopted the self-avoiding random walk as the reference dynamics.

\emph{\textbf{Symmetry} ($h = \{2,3,4\} $)}: This measurement~\cite{silva2016concentric}, obtained for each node $i$, quantifies the symmetry of the topology around $i$. It can be understood as a normalization of the accessibility, and includes two components: \emph{backbone}, where edges between nodes from the same concentric level are discarded, and \emph{merged}, where nodes that share edges in the same level are merged. 

Network visualization can provide means to better understand the structure of a given book's story by organizing, into an embedding space, the topology of the obtained network.  We applied a visualization methodology based on force-directed graph drawing~\cite{silva2016using}. Specifically, this method is based on the \citet{fruchterman1991graph} (FR) algorithm, which simulates a system of particles, which attract and repel one another. The attractive force, $f_a$, reflects the node connectivity, while the repulsive force, $f_r$, acts between all pair of nodes. A gravitational force, $f_g$, can also be added. We adopted $f_a = 0.0002$, $f_r = 1.25$, and $f_g = 0.001$.

\subsection{Machine Learning Methods}

Several classifiers ---  Decision Trees, Random Forest, kNN, Logistic Regressors, SVM, Naive Bayes~\cite{Duda} --- were tested in order to choose the most adequate.  Support Vector Machines (SVM) and Random Forest were selected. We used the Linear SVM implementation (with default parameters), and Random Forest with 50 trees, both available at \emph{Scikit-learn}~\cite{pedregosa2011scikit}. We employed the \emph{leave-one-out} cross-validation technique, in which only one dataset instance is used as test while all the others are taken for training the classifier. Feature selection was attempted, but no particular subset of features stood out.  Therefore, all measurements were considered.

\section{Results and Discussion}\label{sec:results}
In this section, we describe the selected dataset and present the obtained results organized in two parts: (i) the complete set of authors; and (ii) four authors representing major types of works.

\subsection{Dataset}
In order to investigate whether authors can be distinguished by the story flow in their works, we created mesoscopic networks from several texts. Our dataset is composed of 100 English texts written by 20 distinct authors (five texts per author) extracted from~\citet{Machicao}. The selected 20 authors are: {Andrew Lang}, {Arthur Conan Doyle}, {B. M. Bower}, {Bram Stoker}, {Charles Darwin}, {Charles Dickens}, {Edgar Allan Poe}, {H. G. Wells}, {Hector H. Munro (Saki)}, {Henry James}, {Herman Melville}, {Horatio Alger}, {Jane Austen}, {Mark Twain}, {Nathaniel Hawthorne}, {P. G. Wodehouse}, {Richard Harding Davis}, {Thomas Hardy}, {Washington Irving}, and {Zane Grey}. The whole dataset was obtained from the Project Gutenberg repository\footnote{Project Gutenberg - https://www.gutenberg.org/}.  The complete list of used texts is presented in Table~\ref{books}.

\subsection{Complete Set of Authors}
In the first experiment, we used all the books by all 20 authors, yielding the results presented in Table~\ref{resultsperavgdegree}. Remarkably, though the chance baseline for this experiment is only 5\% (each author has the same probability of being randomly selected), our best result was as high as 35\%. {Moreover, 17 (48.5\%) out of the 35 books correctly classified by our method were written by only 4 authors: namely {Andrew Lang}, {B. M. Bower}, {Hector H. Munro (Saki)}, and {Henry James}}

\begin{table}[!htpb]
\caption{Accuracy rate in discriminating the authorship of texts.}\label{resultsperavgdegree}
\begin{center}
\begin{tabular}{lll}
%\hline
\bf Average Degree & \bf Random Forest & \bf SVM\\ 
\hline
$\left\langle k \right \rangle=5$ & $10\%$ & $12\%$\\
$\left\langle k \right \rangle= 10$ & $18\%$ & $14\%$\\
$\left\langle k \right \rangle= 15$ & $22\%$ & $25\%$\\
$\left\langle k \right \rangle= 20$ & $25\%$ & $24\%$\\
$\left\langle k \right \rangle= 25$ & $21\%$ & $17\%$\\
$\left\langle k \right \rangle= 30$ & $21\%$ & $23\%$\\
$\left\langle k \right \rangle= 35$ & $16\%$ & $17\%$\\
$\left\langle k \right \rangle=40$ & $16\%$ & $23\%$\\
$\left\langle k \right \rangle= 45$ & $18\%$ & $25\%$\\
$\left\langle k \right \rangle=50$ & $16\%$ & $20\%$\\
\hline
All combined & $26\%$ & \textbf{35\%}\\
\end{tabular}
\end{center}
\end{table}

\begin{figure*}[!htpb]
 \centering
   \includegraphics[height=0.68 \linewidth]{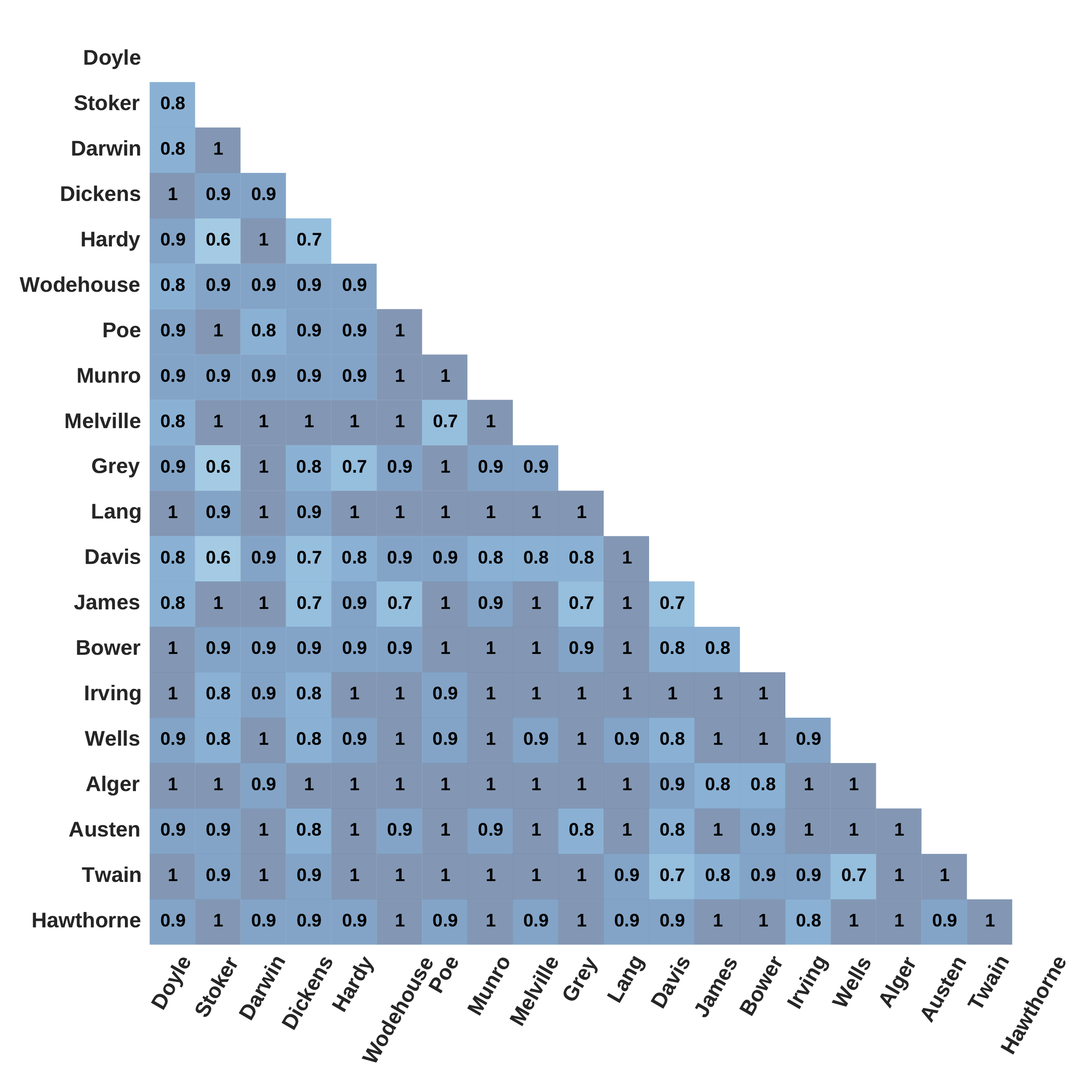}
    \caption{Accuracy rate (from 0 to 1) in the pairwise classification using the frequency of the 20 most frequent words.}
\label{fig:most_frequent_results}
\end{figure*}

\begin{figure*}[!htpb]
 \centering
   \includegraphics[height=0.68 \linewidth]{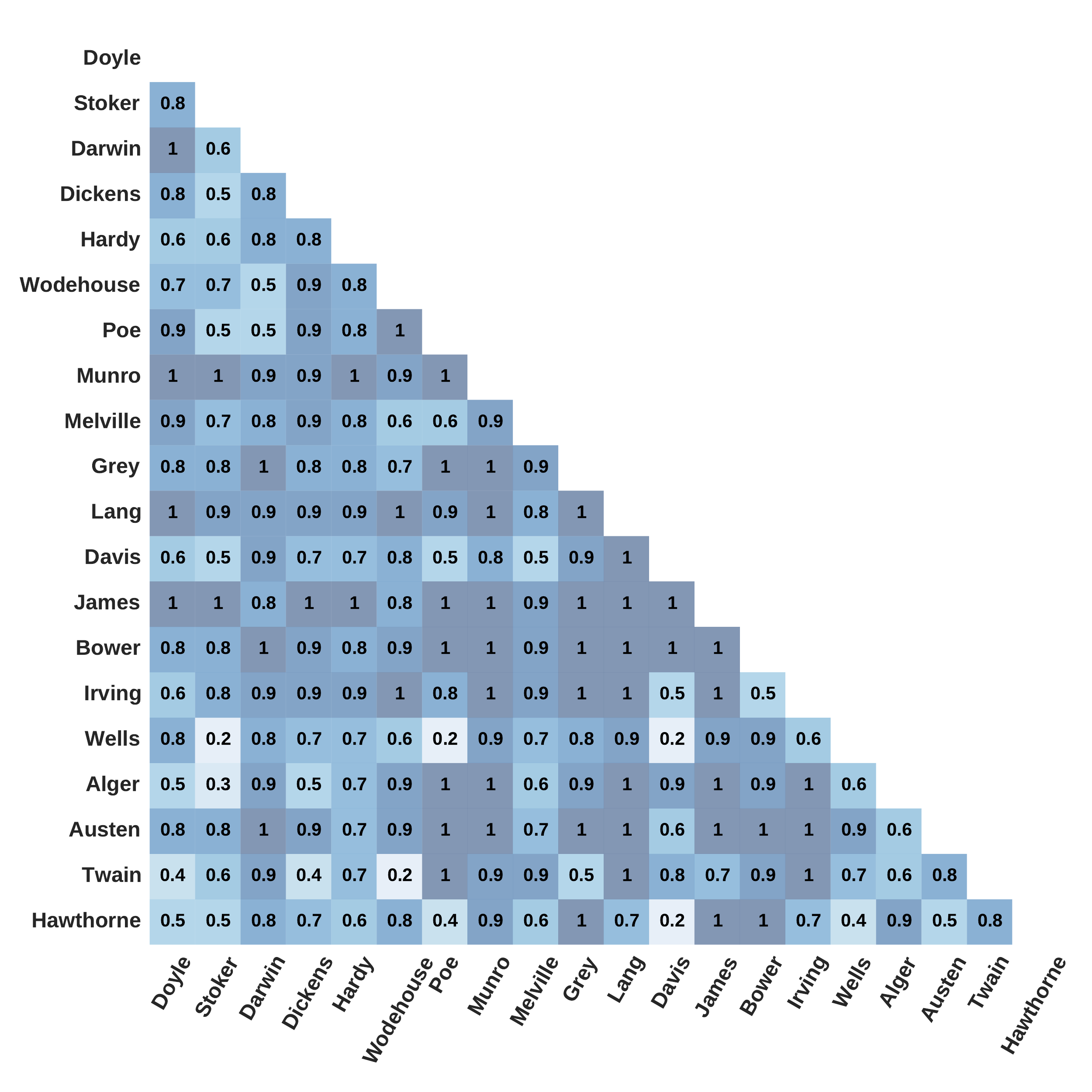}
    \caption{Accuracy rate (from 0 to 1) in the pairwise classification using network features extracted from mesoscopic networks.}
\label{fig:mesoscopic_results}
\end{figure*}

We also performed a pairwise classification. The obtained results were compared with {a traditional approach usually employed in the literature, the analysis of the most frequent words. For this experiment, we used the original texts of each book, extracted the frequency of the 20 most frequent words, and then used a SVM classifier.} Figure~\ref{fig:most_frequent_results} shows the accuracies for the traditional features, and Figure~\ref{fig:mesoscopic_results} illustrates the pairwise classification accuracies when mesoscopic networks were used to model each text, {we did not select a single average degree $\left\langle k \right \rangle$, but rather we combined all the degrees listed in Table~\ref{resultsperavgdegree}. The accuracies were obtained with the SVM classifier}.

A careful examination of Figure~\ref{fig:most_frequent_results} and~\ref{fig:mesoscopic_results} reveals that for some cases, except the squares with lighter colors, our results are on par with those obtained with the frequency of the 20 most frequent words (mainly stopwords). Moreover, our method even achieved higher accuracies in some combinations. See, for example, authors Grey and Munro, for which 7 and 6, respectively, of our results were better than the traditional approach. One thing that we should note, and which will be revisited in the following subsection, is the fact that it is hard for mesoscopic networks to distinguish Edgar Allan Poe from Charles Darwin.  In this case, we obtained an accuracy rate of 50\%, contrasted to 80\% achieved by the other approach.

\subsection{Small Set of Authors}
\begin{figure}[!htpb]
 \centering
   \includegraphics[width=1.0 \linewidth]{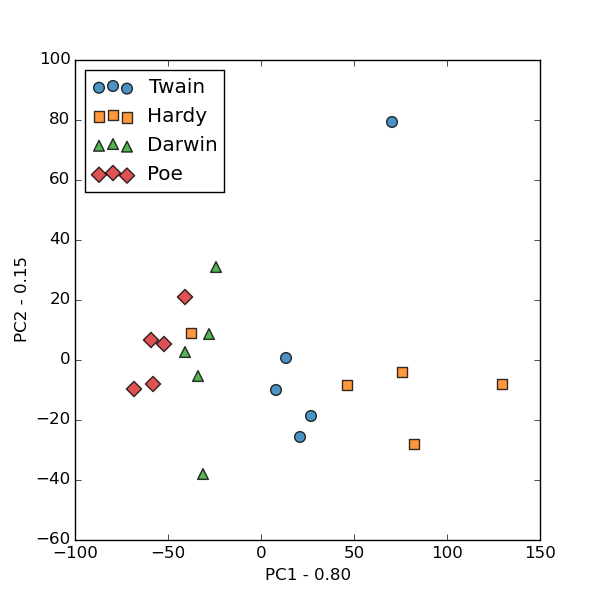}
    \caption{PCA of the books written by Charles Darwin, Thomas Hardy, Edgar Allan Poe, and Mark Twain.}
 \label{fig:pca}
\end{figure}

Out of the 20 authors considered in the previous subsection, we selected four authors, namely Charles Darwin, Thomas Hardy, Edgar Allan Poe, and Mark Twain. They were chosen because two of them have several \emph{novels} (Thomas Hardy and Mark Twain), Edgar Allan Poe is best known for writing \emph{short stories} and Charles Darwin wrote about his \emph{scientific theories} and observations. The now obtained accuracy rate in classifying them was enhanced to 65\% (Random Forests) and 50\% (SVM) by using the mesoscopic representation, contrasted to the chance baseline of 25\% obtained for four authors. The Principal Component Analysis (PCA)~\cite{jolliffe2002principal} considering these four authors is presented in Figure~\ref{fig:pca}.

\begin{figure*}
 \centering
   \includegraphics[width=1.0 \linewidth]{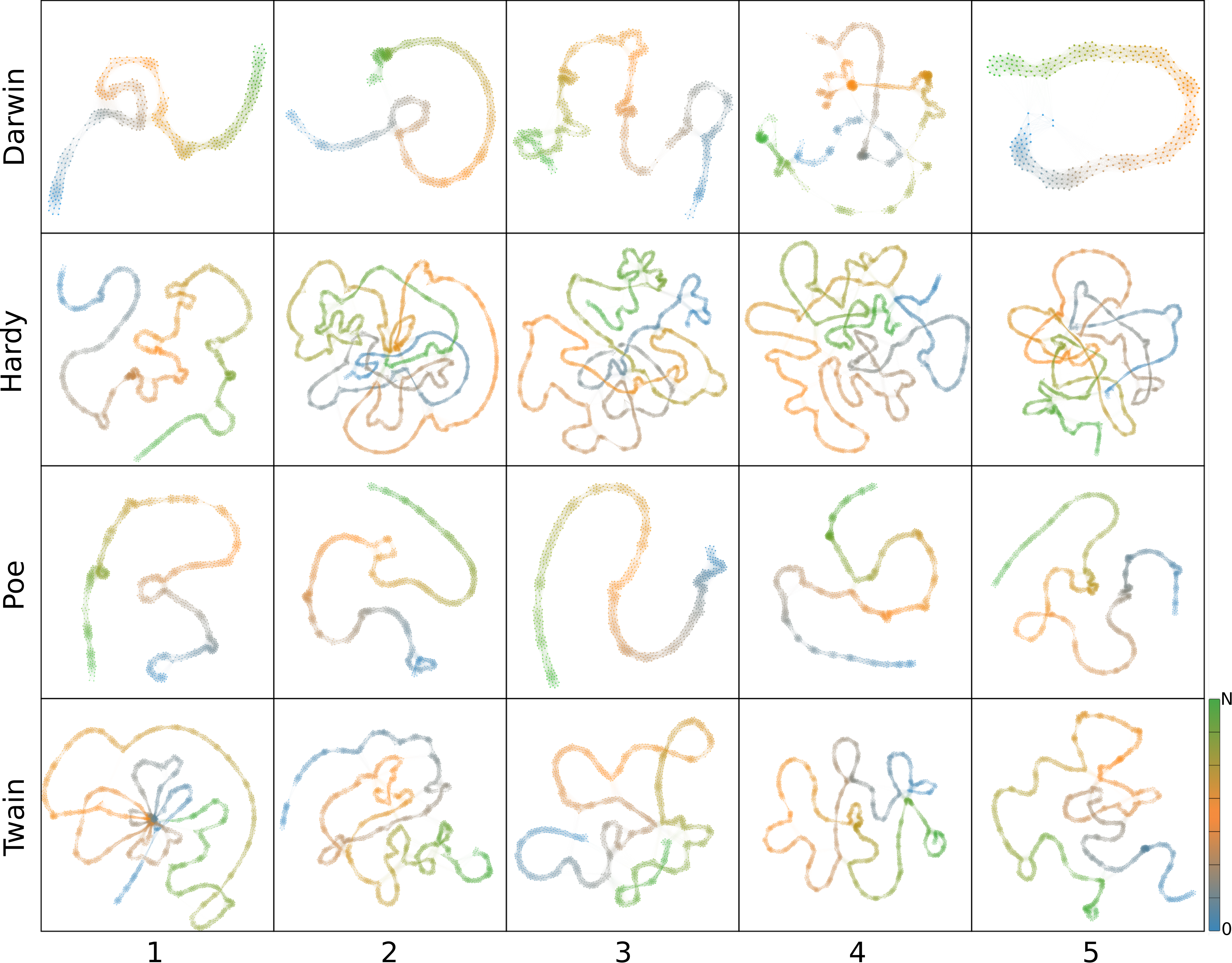}
    \caption{Mesoscopic networks for 20 books of four different authors. \textbf{Charles Darwin:} (1) \emph{Coral Reefs}, (2) \emph{The Expression of the Emotions in Man and Animals}, (3) \emph{Geological Observations on South America}, (4) \emph{The Different Forms of Flowers on Plants of the Same Species}, and (5) \emph{Volcanic Islands}. \textbf{Thomas Hardy:} (1) \emph{A Changed Man; and Other Tales}, (2) \emph{A Pair of Blue Eyes}, (3) \emph{Far from the Madding Crowd}, (4) \emph{Jude the Obscure}, and (5) \emph{The Hand of Ethelberta}. \textbf{Edgar Allan Poe:}  \emph{The Works of Edgar Allan Poe - Volume (1) to (5)}. \textbf{Mark Twain:} (1) \emph{Adventures of Huckleberry Finn}, (2) \emph{The Adventures of Tom Sawyer}, (3) \emph{The Prince and the Pauper}, (4) \emph{A Connecticut Yankee in King Arthur's Court}, and (5) \emph{Roughing It}. The bluish nodes represent the windows formed by paragraphs from the beginning of the book and the greenish ones represent the windows formed by paragraphs from the end of the book. The order of the windows can be seen in the legend, where $N$ represents the last window.}
 \label{redes_4}
\end{figure*}

The PCA results indicate a clear partitioning between the groups of books associated to each author. Remarkably, one of Thomas Hardy's book (\emph{A Changed Man and Other Tales}) resulted between those of Edgar Allan Poe and Charles Darwin. Such a good partitioning is a consequence of the quite different mesoscopic networks obtained for these authors, as depicted in Figure~\ref{redes_4}. 

The mesoscopic networks presented in Figure~\ref{redes_4} unveil interesting aspects, including an unexpected similarity to intricate calligraphic shapes. Note that the books which contain tales or short stories, such as those by Edgar Allan Poe, as well as the book \emph{A Changed Man and Other Tales}, present a similar chain-like topology with a few cycles. Moreover, most of these cycles appear at a relatively small scale. Interestingly, the scientific books of Charles Darwin also present this chain-like structure, which is probably related to the nature of his writings, describing his theories, observations, and findings.

It is clear, visually, that the other books present more complex stories, where paragraphs (nodes) from different parts of the book sharing similar content resulted in intersections. For example, the book \emph{Adventures of Huckleberry Finn} tells the story of Huckleberry Finn traveling down the Mississippi river. During most of the book, he goes through different small adventures along the river. Another interesting point is that this book ends in a similar setting as it begins, when Huckleberry Finn returns to his city, which is reflected in the respective return of the unfolding trajectory to its beginning. {It is important to highlight that a full visual analysis with all the 20 authors was beyond the scope of this experiment. Our primary goal was to perform a preliminary investigation of the books through geometrical approaches.}

\section{Conclusion}\label{sec:conclusion}
Complex network methods have been applied with growing success to several natural language processing tasks. In some of these approaches, a chunk of text is represented as a co-occurrence network, which reflects the syntactic relationship between words~\cite{Cancho01thesmall}. Although this is a well-known representation, it is not without its share of problems. Those networks, for example, are unable to represent the topical structure found in many texts. So as to overcome such a limitation, a mesoscopic representation has been recently proposed \cite{de2017mesoscopic}. The main goal of that approach was to take into account the semantical relationship between chunks of text. More specifically, the network nodes correspond to texts from consecutive paragraphs, while the edges are weighted by the similarity between the respective texts. Statistics of some local topological measurements were used to characterize books' mesoscopic networks. We tested the hypothesis that such a representation is useful at assigning the authorship to documents. In particular, we advocated that fingerprints left by each author are visible at a mesoscopic scale.

The obtained accuracy rates, which in one case surpassed by 40 percentage points the chance baseline, suggest that the proposed approach is capable of revealing writing styles characteristics. In addition, we performed an alternative classification, in which all pairs of distinct authors were considered. In some cases our method provided better results than those obtained with traditional features. Such a result indicates that features obtained from mesoscopic networks can be used as a complement to more traditional features of texts. In order to better understand the unfolding of texts, we selected authors whose works include short stories, novels, and scientific writing. A set of topological features was estimated and PCA projected.  Interestingly, in this projected space, a book of tales written by \emph{Thomas Hardy} resulted closer to \emph{Edgar Allan Poe}'s books, which are also composed of short stories. Even more surprising, the patterns obtained by the visualization resulted quite representative of the different types of works, suggesting a ``calligraphy''. Such visualizations reveal intricate discourse patterns in the books.

{The goal of this paper was not to provide state-of-the-art results for authorship attribution, given that most traditional approaches in the literature have achieved results as high as 90\%~\cite{grieve2007,Koppel:2009}. Instead, we report an approach that can be used to obtain novel stylometric features, as well as to complement traditional methods.}

Future works could apply a similar approach to other related tasks --- such as authorship verification, plagiarism detection, and topic segmentation ---  and also extend the mesoscopic representation to include different granularity levels, such as sentences or chapters. Another possibility is to investigate the relationship between the emotional content of a text and its topology.

\section*{Acknowledgments}
V.Q.M. and D.R.A. acknowledge financial support from S\~ao Paulo Research Foundation (FAPESP) (grant no. 15/05676-8, 16/19069-9). 
H.F.A. and T.S.L. thank CAPES for financial support. L.d.F.C. is grateful to CNPq (Brazil) (grant no.  307333/2013-2), FAPESP (grant no.  11/50761-2), and NAP-PRP-USP for sponsorship.

%\bibliography{acl2017}
\bibliography{acl2017}
\bibliographystyle{acl_natbib}

\appendix

\section{Dataset}

\begin{table*}
\caption{List with the 100 texts employed in the authorship attribution task.}
\label{books}
\centering
\begin{tabular}{p{8cm}||p{8cm}}
\hline
 \textbf{Author: Texts} &  \textbf{Author: Texts}\\
\hline
\textbf{Andrew Lang:} The Arabian Nights Entertainments; The Blue Fairy Book; The Pink Fairy Book; The Violet Fairy Book; The Yellow Fairy Book & \textbf{Herman Melville:} Moby Dick, Or, The Whale; The Confidence-Man: His Masquerade; The Piazza Tales; Typee: A Romance of the South Seas; White Jacket, Or, The World on a Man-of-War\\
\hline
\textbf{Arthur Conan Doyle:} The Tragedy of the Korosko; The Valley of Fear; The War in South Africa; Through the Magic Door; Uncle Bernac - A Memory of the Empire & \textbf{Horatio Alger:} Adrift in New York: Tom and Florence Braving the World; Brave and Bold, Or, The Fortunes of Robert Rushton; Fame and Fortune or, The Progress of Richard Hunter; Ragged Dick, Or, Street Life in New York with the Boot-Blacks; The Errand Boy, Or, How Phil Brent Won Success\\
\hline
\textbf{B. M. Bower:} Cabin Fever; Lonesome Land; The Long Shadow; The Lookout Man; The Trail of the White Mule & \textbf{Jane Austen:} Emma; Mansfield Park; Persuasion; Pride and Prejudice; Sense and Sensibility\\
\hline
\textbf{Bram Stoker:} Dracula's Guest; Lair of the White Worm; The Jewel Of Seven Stars; The Lady of the Shroud; The Man & \textbf{Mark Twain:} A Connecticut Yankee in King Arthur's Court; Adventures of Huckleberry Finn; The Adventures of Tom Sawyer; The Prince and the Pauper; Roughing It\\
\hline
\textbf{Charles Darwin:} Coral Reefs; Geological Observations on South America; The Different Forms of Flowers on Plants of the Same Species; The Expression of the Emotions in Man and Animals; Volcanic Islands & \textbf{Nathaniel Hawthorne:} Mosses from an Old Manse, and Other Stories; The Blithedale Romance; The House of the Seven Gables; The Scarlet Letter; Twice Told Tales\\
\hline
\textbf{Charles Dickens:} American Notes; A Tale of Two Cities; Barnaby Rudge: A Tale of the Riots of Eighty; Great Expectations; Hard Times & \textbf{P. G. Wodehouse:} My Man Jeeves; Tales of St. Austin’s; The Adventures of Sally; The Clicking of Cuthbert; The Man with Two Left Feet\\
\hline
\textbf{Edgar Allan Poe:} The Works of Edgar Allan Poe (Volume 1 - 5) & \textbf{Richard Harding Davis:} Cinderella, and Other Stories; Notes of a War Correspondent; Real Soldiers of Fortune; Soldiers of Fortune; The Congo and Coasts of Africa\\
\hline
\textbf{Hector H. Munro (Saki):} Beasts and Super Beasts; The Chronicles of Clovis; The Toys of Peace; The Unbearable Bassington; When William Came & \textbf{Thomas Hardy:} A Changed Man and Other Tales; A Pair of Blue Eyes; Far from the Madding Crowd; Jude the Obscure; The Hand of Ethelberta\\
\hline
\textbf{Henry James:} The Ambassadors; The American; The Portrait of a Lady - Volume 1; The Real Thing and Other Tales; The Turn of the Screw & \textbf{Washington Irving:} Chronicle of the Conquest of Granada, from the mss. of Fray Antonio Agapida; Knickerbocker’s History of New York; Tales of a Traveller; The Alhambra; The Sketch-Book of Geoffrey Crayon\\
\hline
\textbf{H. G. Wells:} A Short History of the World; Tales of Space and Time; The First Men in the Moon; The War of the Worlds; The World Set Free & \textbf{Zane Grey:} Riders of the Purple Sage; The Call of the Canyon; The Lone Star Ranger: A Romance of the Border; The Mysterious Rider; To the Last Man\\
\hline
\end{tabular}
\end{table*}

\end{document}